\documentclass[preprint,12pt,authoryear]{elsarticle}

\usepackage{amsmath,amsfonts,bm}

\def\eqref#1{equation~\ref{#1}}

\def\1{\bm{1}}

\DeclareMathAlphabet{\mathsfit}{\encodingdefault}{\sfdefault}{m}{sl}
\SetMathAlphabet{\mathsfit}{bold}{\encodingdefault}{\sfdefault}{bx}{n}

\usepackage{amssymb}
\usepackage{hyperref}
\usepackage{url}

\usepackage{bm}
\usepackage{graphicx}
\usepackage{wrapfig}
\usepackage{xcolor}
\usepackage[utf8]{inputenc} 
\usepackage[T1]{fontenc}    
\usepackage{hyperref}       
\usepackage{url}            
\usepackage{booktabs}       
\usepackage{amsfonts}       
\usepackage{nicefrac}       
\usepackage{microtype}      

\usepackage{graphicx}    %
\usepackage{subfig}
\graphicspath {{images/}{images/occlusion/}}

\usepackage{tabu}
\usepackage[inline]{enumitem}
\usepackage{multirow}
\usepackage{caption}
\usepackage{multicol}
\usepackage{makecell}

\usepackage{color}

\journal{Neural Networks}

\begin{document}

\begin{frontmatter}



\title{TraceCaps: A Capsule-based Neural Network for Semantic Segmentation}

\author[OUaddress]{ Tao Sun}
\author[OUaddress]{ Zhewei Wang}
\author[UKaddress]{ Charles D. Smith}
\author[OUaddress]{ Jundong Liu\corref{correspondingauthor}}
\cortext[correspondingauthor]{Corresponding author. Email: liuj1@ohio.edu}

\author[]{For the Alzheimer's Disease Neuroimaging
  Initiative\corref{ADNI_data}} \cortext[ADNI_data]{Data used in
  preparation of this article were obtained from the Alzheimer’s
  Disease Neuroimaging Initiative (ADNI) database
  (\url{adni.loni.usc.edu}). As such, the investigators within the
  ADNI contributed to the design and implementation of ADNI and/or
  provided data but did not participate in analysis or writing of this
  report. A complete listing of ADNI investigators can be found at:
  \url{http://adni.loni.usc.edu/wp-content/uploads/how_to_apply/ADNI_Acknowledgement_List.pdf}}

\address[OUaddress]{School of Electrical Engineering and Computer
  Science \\ Ohio University, Athens OH 45703}
\address[UKaddress]{Department of Neurology \\ University of Kentucky,
  Lexington KY 40503}

\begin{abstract}
  In this paper, we propose a capsule-based neural network model to
  solve the semantic segmentation problem. By taking advantage of the
  extractable part-whole dependencies available in capsule layers, we
  derive the probabilities of the class labels for individual capsules
  through a recursive, layer-by-layer procedure. We model this
  procedure as a {\it traceback pipeline} and take it as a central
  piece to build an end-to-end segmentation network. Under the
  proposed framework, image-level class labels and object boundaries
  are jointly sought in an explicit manner, which poses a significant
  advantage over the state-of-the-art fully convolutional network
  (FCN) solutions.  With the capability to extracted part-whole
  information, our {\it traceback pipeline} can potentially be
  utilized as the building blocks to design interpretable neural
  networks.  Experiments conducted on modified MNIST and neuroimages
  demonstrate that our model considerably enhance the segmentation
  performance compared to the leading FCN variants.
\end{abstract}




\end{frontmatter}


\section{Introduction}

An effective segmentation solution should have a well-equipped
mechanism to capture both semantic (i.e., \textit{what}) and location
(i.e., \textit{where}) information. The fully convolutional network
(FCN) \citep{long2015fully} and its variants
\citep{ronneberger2015unet,noh2015deconv,segnet} constitute a popular
class of solutions for this task, producing state-of-the-art results
in a variety of applications. FCN and its variants (FCNs) are commonly
constructed with an encoder-decoder architecture. In the encoding
path, input images are processed through a number of “convolution +
pooling” layers to generate high-level latent features, which are then
progressively upsampled in the decoder to reconstruct the target pixel
labels. The feature maps produced in higher (coarser) layers and those
in lower (finer) layers contain complementary information: the former
is richer in semantics, while the latter carries more spatial details
that define class boundaries.

Originated from and constructed upon convolutional neural networks
(CNNs) \citep{krizhevsky2012alexnet,simonyan2014vgg}, FCNs’ encoders
inherit some common drawbacks of CNNs, one of which is the lack of an
internal mechanism in achieving viewpoint-invariant
recognition. Traditional CNNs, as well as FCNs, rely on convolution
operations to capture various visual patterns, and
utilize poolings to enable multi-scale processing of the input
images. Rotation invariance, however, is not readily available in both
models. As a result, more data samples or additional network setups
\citep{cohen2016group,cohen2018spherical} would be required for
objects from different viewpoints to be correctly
recognized. The absence of explicit part-whole relationships among
objects imposes another limitation for FCNs -- without such a
mechanism, the rich semantic information residing in the higher layers
and the precise boundary information in the lower layers can only be
integrated in an implicit manner \citep{zhang2018exfuse}.

Capsule nets \citep{hinton2017nips,hinton2018ICLR,hinton2019Nips},
operating on a different paradigm, can provide a remedy. Capsule nets
are built on capsules, each of which is a group of neurons
representing one instance of a visual entity, i.e., an object or one
of its parts \citep{hinton2011autoencoder}. Capsules output both
activation probabilities of their presence and the instantiation
parameters that describe their properties, such as pose, deformation
and texture, relative to a viewer \citep{hinton2011autoencoder}.
{\color{black} So far, two types of capsule nets have been proposed,
  following supervised learning \citep{hinton2017nips,hinton2018ICLR}
  and unsupervised learning \citep{hinton2019Nips} paradigms,
  respectively. Our proposed model in this work is developed along the
  line of supervised models, and therefore we will base our discussion
  mostly on supervised capsule nets, unless we point out otherwise.}
During inference propagation, capsule nets
\citep{hinton2017nips,hinton2018ICLR} rely on the principle of
coincidence filtering to activate higher-level capsules and set up
part-whole relationships among capsule entities. Such part-whole
hierarchy equips capsule nets with a solid foundation for
viewpoint-invariant recognition, which can be implemented through
dynamic routing \citep{hinton2017nips} or EM routing
\citep{hinton2018ICLR}. The same hierarchy, if properly embedded into
a segmentation network, would provide a well-grounded platform to
specify contextual constraints and enforce label consistency.

With this thought, we develop a capsule-based semantic segmentation
solution in this paper. Our \color{black}{model is built upon
  supervised capsule nets and treat them as} probabilistic graphical
  models capable of inferring probabilistic dependences among visual
  entities, through which part-whole relationships can be explicitly
  constructed. As a concrete implementation, we propose a new
  operation sequence, which we call \textit{traceback pipeline}, to
  capture such part-whole information through a recursive procedure to
  derive the class memberships for individual pixels. We term our
  model \textit{TraceCaps}.

The contributions of our TraceCaps can be summarized as:
\begin{enumerate}
\item In TraceCaps, the class labels for individual spatial
  coordinates within each capsule layer are analytically derived. The
  traceback pipeline in our model, taking advantage of the graphical
  properties of capsule nets, is mathematically rigorous. To the best
  of our knowledge, this is the first work to explore a capsule
  traceback approach for image segmentation. In addition, probability
  maps at each capsule layer are readily available, which makes it
  convenient to conduct feature visualization and layer
  interpretation.
\item In parallel with segmentation, TraceCaps carries out explicit
  class recognition at the same time. Such explicitness poses a
  powerful practical advantage over FCNs.
\item The traceback pipeline is designed under a
  general context, making it applicable to many other potential
  tasks, including object localization and detection,
  action localization and network interpretation.
\end{enumerate}

\section{Background}

A capsule is aimed to represent
an instance of a visual entity, and its outputs include two parts: 1)
the probability of the existence of this instance and 2) instantiation
parameters, which carries certain visual properties of
  the instance. Built on capsules, capsule nets are designed to
overcome the drawbacks of CNNs, in particular, the inability of
handling viewpoint changes and the absence of part-whole relationships
among visual entities.

\begin{figure*} [bt]
\begin{center}
  \begin{tabular}{cc}
      \includegraphics[width=0.9\linewidth]{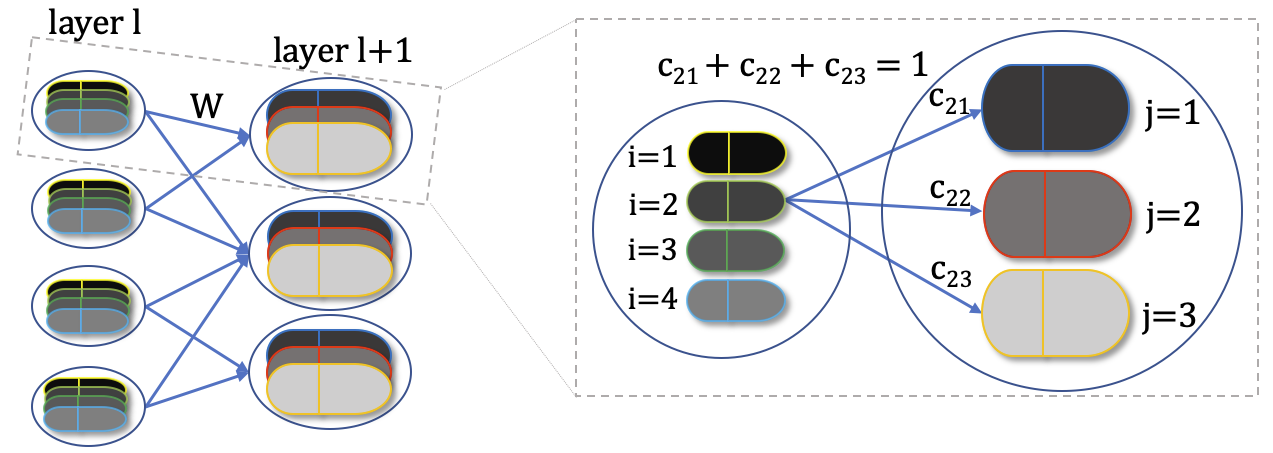}
\end{tabular}
\end{center}
%
\caption{Left diagram: spatial relationships between
  two contiguous capsule layers (shown in 1D for clarity). If a
  position is in the receptive field of another position in the higher
  layer, the two positions are connected with an arrow pointing to the
  higher layer. Each capsule at the higher-layer position is a
  possible parent of capsules at its connected lower-layer
  position. Right diagram: the assignment
  probabilities and its normalization requirement. Refer to text for
  more details.}
  \label{fig:layers}
\end{figure*}

Unlike CNNs, where image patterns are captured through convolution
(correlation) operations, capsule nets
\citep{hinton2017nips,hinton2018ICLR} rely on high-dimensional
coincidence filtering (HDCF) to detect objects that attract
concentrated votes from their parts/children. The HDCF mechanism is
carried out based on the computation of two network parameter sets:
weights and coupling coefficients, of the connected capsules in
consecutive layers (Fig. \ref{fig:layers}).

In CNNs, the weights between neurons represent convolution filters,
which are to be learned in training. In capsule nets, each weight
matrix $W$, however, represents a linear transformation that would map
the parts of an object into a cluster for the same whole. Similar to
CNN, weight matrices in capsule nets are globally learned in training
through labeled samples. As viewpoint changes do not alter the
relative orientation between parts and whole, maintaining the same $W$
would be sufficient to handle inputs from different angles, leading a
viewpoint-invariant system. The coupling coefficients $c_{ij}$ are
assignment probabilities between a pair of whole-part
capsules. Different from the weights, $c_{ij}$ is dynamically
determined solely in inference time. The $c_{ij}$s between each
capsule $i$ and all its potential parent capsules are summed to $1$,
and thus have a similar effect of average pooling in CNNs.

While the two versions of capsule nets
\citep{hinton2017nips,hinton2018ICLR} follow the same HDCF general
principle for object recognition, they differ in several aspects,
primarily the setup of the instantiation parameter $u$ and the routing
process to estimate $c_{ij}$. In \citep{hinton2017nips}, $u$ is set as
a one-dimensional vector, whose length represents the probability of
the capsule's presence. The routing algorithm, {\it dynamic routing},
updates $c_{ij}$ indirectly by the scalar product of votes and outputs
of possible parents. Meanwhile, \cite{hinton2018ICLR} uses a $4 \times
4$ matrix to represent instantiation parameters of a capsule, and
formulates the routing procedure as a mixture of Gaussians problem,
which is then solved with a modified Expectation-Maximization
algorithm.

In this work, we take advantage of the part-whole relationships
available in capsule nets to produce the labels for the pixels at the
input image space. To facilitate the derivation of our solution, we
would first summarize the related concepts under a probabilistic
framework.

Capsules can be categorized into different types (\textit{e.g.} cat,
bicycle, or sofa). Each capsule layer $l$ is associated with a set of
capsule types
$\mathcal{T}^l = \{t_1^{l}, t_2^{l}, ...,t_ m^{l}, ...\}$. At each
spatial coordinate (position) of certain layer $l$, there exists
exactly one capsule of every type in $\mathcal{T}^l$. The probability
of a capsule $i$ existing/activated in the network is denoted as
$P(i)$. If capsule $i$ on layer $l$ is within the receptive field of
the capsule $j$ of layer $l+1$, we call $j$ a possible parent of
$i$. The assignment probability of $i$ being a part of $j$ is denoted
as $P(j|i) = c_{ij}$. The assignment probabilities between capsule $i$
and the capsules co-located with $j$ are forced to meet the
normalization requirements of
$\sum_{j \in \mathcal{T}^{l+1}} c_{ij}=1$. 

Three types of capsule layers were introduced in
\citep{hinton2017nips,hinton2018ICLR}, which will also be used as
building blocks in our TraceCaps. Therefore, we briefly describe them
as follows,
\begin{itemize}
\item Primary capsule layer is the first capsules layer, where
  features from previous convolution layer are processed and
  transitioned into capsules via convolution
    filtering.
\item Convolutional capsule layer(s) function
    similarly to CNN convolutional layers in many aspects.  However,
    they take capsules as inputs and utilize certain routing algorithm
    to infer outputs, which are also capsules.
\item Class capsule layer $L$ is a degenerated layer with one capsule
  for each predefined class label
  $\mathcal{C}_k \in \mathcal{C} = \{\mathcal{C} _1, \mathcal{C} _2,
  ...,\mathcal{C} _k, ...\}$. Each capsule in the previous layer is
  fully connected to the capsules in this layer.
\end{itemize}

Capsule nets are designed for object classification. In this work, we
aim to take advantage of the notion and structure of capsule nets to
design a semantic segmentation solution, hoping the viewpoint
invariance and network interoperability mechanism can lead to improved
segmentation performance. We present the description of our model in
next section.

\section{The Architecture of Our TraceCaps}

\begin{figure*} [bt]
  \centering
      \includegraphics[width=0.95\linewidth]{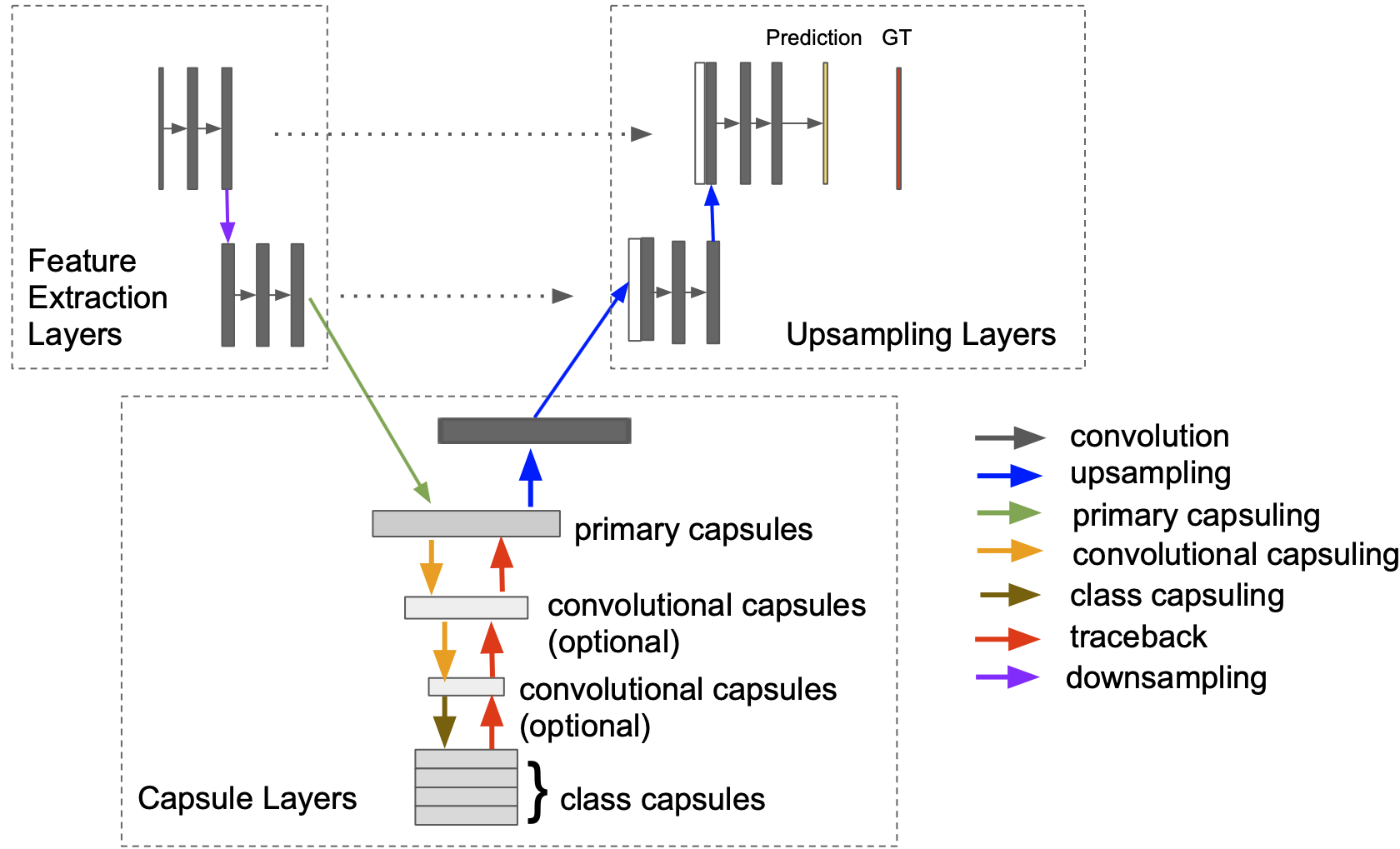}
 
      \caption{Overall architecture of our
          TraceCaps. The traceback pipeline, shown as red arrows, is
          the major innovation of this paper. {\it GT} stands for
          ground-truth. Refer to text for details.}
  \label{fig:arch}
\end{figure*}

The general structure of our TraceCaps, which consists of three
components, is shown in Fig.~\ref{fig:arch}. It starts with a
\textit{feature extraction module} that capture the
discriminative features of the input data to be fed into the later
modules. For this purpose, a sequence
  of convolution layers as in CNNs would be a good choice
\citep{hinton2017nips,hinton2018ICLR}. A
\textit{capsule \& traceback module} comes next.
It consists of a sequence of capsule layers and a
  traceback pipeline, followed by a convolution layer.  The lineup of
  the capsule layers, shown as orange and brown arrows in
  Fig.~\ref{fig:arch}, can be a combination of one primary capsule
  layer, optionally one or more convolutional capsule layers, and a
  class capsule layer at the end. The traceback pipeline, shown as red
  arrows in Fig.~\ref{fig:arch}, is the major innovation of this
  paper.  It produces class maps of the same size as the primary
  capsule layer, which are then taken as inputs to a convolution
  layer. This pipeline is specifically designed for segmentation
  purpose, and its details will be presented in next subsection.
The third component of our model, a \textit{upsampling
    module}, restores the original resolution by upsampling the label
  map computed in the capsule \& trackback module. This module is
  implemented based on the deconvolution scheme in
  \citep{long2015fully}, but it can also be implemented
  through atrous convolution \citep{chen2018deeplab}, unpooling
  \citep{segnet}, or simply bilinear interpolation. It should be
noted that the feature extraction and upsampling layers can be
regarded as somewhat symmetric in terms dimensionality, where the
latter can be optional if no dimensionality reduction occurs in
feature extraction layers.

\subsection{Traceback pipeline: design and derivations}		

The ultimate goal of image segmentation is to compute the probability
of each pixel belonging to certain class type, hopefully in great
accuracy.  The traceback pipeline is designed to serve this
purpose. It should be noted that, over the inference procedure of 
capsule nets, the probability of each capsule $P(i)$ and the
assignment probabilities between contiguous layers of capsules
$c_{ij}$ are calculated and become available. With that,
$P(\mathcal{C}_{k})$, the probability of a class label for each
location in the capsule layers, can be potentially inferred through
repeated applications of the product rule and the sum rule in
probability theory. If this process is carried out, it would
essentially trace the class labels in the class capsule layer in a
backward-propagation manner, layer by layer until it reaches the first
capsule layer. We name this recursive, layer-by-layer
  process \textit{traceback pipeline}. \cite{hinton2000learning}
takes a similar approach, which interprets an image as a parse tree to
perform recognition and segmentation simultaneously. Their model was
constructed based on graphical models though, with no capsule or other
neural network involved.

The detailed traceback procedure is explained as follows. The feature
extraction layers and the capsules layers of TraceCaps are adopted
from the capsule nets. Therefore, same as in the latter, $P(i)$ and
$c_{ij}$ are available during the inference procedure of
TraceCaps. With that, the probability of a position belonging to
certain class $P(\mathcal{C}_{k})$, in layer $l$, can be calculated as

\begin{equation} \label{eq:segclass} P(\mathcal{C}_{k}) = \sum_{i \in
    \mathcal{T}^l} P(\mathcal{C}_{k}, i) = \sum_{i \in \mathcal{T}^l}
  P(i)P(\mathcal{C}_{k} | i) ,
\end{equation}

\noindent where $i$ is a capsule type associated with layer $l$ and
$P(\mathcal{C}_{k} | i)$ shows the likelihood of certain position
taking $\mathcal{C}_{k}$ as its class label, given that a capsule with
type $i$ is located on it. With $P(i)$ of layer $l$ being available
over the inference procedure, $P(\mathcal{C}_{k} | i)$ is the only
term to be estimated.

Let $L$ be the index of the class capsule layer. The
$P(\mathcal{C}_{k} | i)$ at its immediately previous layer, $L-1$,
would be the assignment probability between the capsule $i$ of the
layer $L-1$ and the class capsule $\mathcal{C}_{k}$ on layer $L$.
Again, $P(\mathcal{C}_{k} | i) $ is available after inference reaches
layer $L$.

The $P(\mathcal{C}_{k} | i) $ on other layers, however, needs to be
solved. After some simple mathematical derivations, we found that the
estimation of $P(\mathcal{C}_{k} | i) $ could be written into a
recursive equation w.r.t. the upper layer, if we assume that each
lower-layer capsule only takes capsules in one particular position of
the higher-layer as its possible parents. Let capsule $j$ of layer
$l+1$ and capsule $i$ of layer $l$ form a possible parent-child
pair. Then the conditional probability $P(\mathcal{C}_{k} | i)$ can be
computed as

\begin{equation} \label{eq:seginducive}
\begin{aligned}
P(\mathcal{C}_{k} | i)  = & \sum_{j \in \mathcal{T}^{l+1}} P(\mathcal{C}_{k}, j | i) \quad  (i \text{ is assigned to the parent } j.) \\ 
  = & \sum_{j \in \mathcal{T}^{l+1}} P(j|i) P(\mathcal{C}_{k} | j, i) \\
  = & \sum_{j \in \mathcal{T}^{l+1}} c_{ij} P(\mathcal{C}_{k} | j) \quad  (i \text{ has the same class as its parent } j.)
\end{aligned}
\end{equation}

\begin{figure*} [bt]
  \centering
  \includegraphics[width=0.65\linewidth]{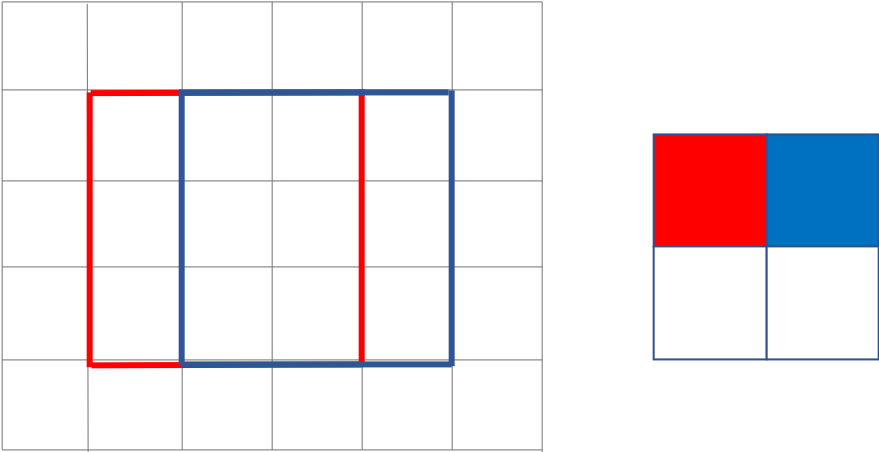}
  \caption{Illustration of a {\it
        parents-at-multi-positions} scenario. Left: a capsule layer,
      where each cell represents a capsule. Right: four capsules in
      the next layer, generated from the previous layer (left) through
      a 3$\times$3 convolution with stride 1. The capsules in the
      overlapping area of the two frames (left) take both the red and
      blue capsules in the higher-level layer (right) as parents.}
  \label{fig:deconv}
\end{figure*}

The recurrence in Eqn. (\ref{eq:seginducive}) indicates that the
estimation of $P(\mathcal{C}_{k} | i)$ requires the conditional
probabilities $P(\mathcal{C}_{k} | j)$ from the parent layer.

We should note that when the inference propagation reaches layer $L$,
all $c_{ij}$ are available. In addition, the $P(\mathcal{C}_{k} | j)$
are also available for layer $L-1$. With this combination, the
$P(\mathcal{C}_{k} | i)$ on other layers can be computed through a
traceback process. More specially, $P(\mathcal{C}_{k} | i) $ can be
estimated with a layer-by-layer backward propagation procedure --
starting at the layer $L - 1$, and repeatedly applying
Eqn. (\ref{eq:seginducive}) to compute the conditional probabilities
for the lower layers. We use a term
  \textit{traceback-depth} to indicate the number of capsule layers to
  be traced along the pipeline. The size of the last capsule layer, of
  which the conditional probabilities are calculated, is termed
  \textit{traceback-size}.

Eqn. (\ref{eq:seginducive}) is for a simple case where each
lower-layer capsule only takes same-position capsules of the
higher-layer as its possible parents. For the convolutional capsule
layers, however, capsules in two or more positions might be the
parents of a lower-layer capsule.
Fig. \ref{fig:deconv} shows an example where a number
  of capsules take capsules at different positions in next layer (blue
  and red) as their parents.  For these cases, the traceback
procedure remains effective, but the computation of
$P(\mathcal{C}_{k} | i)$ needs to be modified. One possible approach
is to stipulate 
$P(\mathcal{C}_{k} | i) = \sum_n P_n(\mathcal{C}_{k} | i) / N$, where
$n$ represents $n$-th location for possible parents of the capsule $i$
and $N$ is total location number. This summation bears some
resemblance to the summation in deconvolution operation
\citep{long2015fully} that adds two overlapping filter responses.

\subsection{Loss Function}
The loss function of our model,
$L= \lambda_{1} L_{\textrm{margin}} + \lambda_{2} L_{\textrm{ce}}$, is
a weighted sum of a margin loss term $L_{\textrm{margin}} $ over the
class capsules
and a pixel-wise softmax cross-entropy $L_{\textrm{ce}}$ over the
final feature map.  $\lambda_{1}$ and $\lambda_{2}$ are the weighting
parameters. 


The margin loss, adopted from \citep{hinton2017nips}, is for
recognition, which strives to increase the probabilities of class
capsules for the corresponding objects existing in the input
image. Let $P(\mathcal{C}_{k})$ denote the probability of the class
capsule k. For each predefined class, there is a contributing term
$L_k$ in the margin loss
\begin{equation}
L_k = T_k \ max(0, m^{+}  - P(\mathcal{C}_{k}))^2 + \lambda_{\textrm{margin}} (1-T_k)\ max(0, P(\mathcal{C}_{k}) - m ^{-})^2
\end{equation}

\noindent
where $m^{+} =0.9$, $m^{-} =0.1$, and
$\lambda_{\textrm{margin}} =0.5$. $T_k=1$ when a class $k$ is present
in the mask and $T_k=0$ otherwise. The margin loss is the summation of
all $L_k$, i.e. $L_{\textrm{margin}} =\sum_{k} L_k$.

The margin loss plays a key role in training the model because it is
the driving force for the initiation and convergence of the capsule
layers in the model. While the cross-entropy loss handles the boundary
localization in semantic segmentation, the margin loss is responsible
for the object recognition. As a result, the ratio
$\frac{\lambda_{1}}{\lambda_{2}}$ in the total loss equation is an
important hyper-parameter to tune the performance of our models, which
we will discuss in next section.

\section{Experimental Results} \label{sec:exp}

\paragraph{Datasets} The effectiveness of our model is evaluated with
two datasets: modified MNIST and Hippocampus dataset from the
Alzheimer’s Disease Neuroimaging Initiative (ADNI) project.

The MNIST \citep{lecun1998mnist} dataset contains 28$\times$28 images
of handwritten digits. In our experiments, random noise is added to
each image to broaden the intensity spectrum and increase the numerical stability. Ground-truth
segmentations are generated by filtering original foreground digit
pixels of each image with an intensity threshold. Two experiments are
conducted on this noise-added MNIST dataset. The first one is a
multi-digit segmentation test, in which competing methods are
evaluated based on their performance in segmenting all ten digits. The
second experiment is a robustness test to assess the models under
various occlusion scenarios.

\begin{figure*} [bt]
  \begin{center}
  \begin{tabular}{cc}
     \multirow{2}{*}[6.5em]{\includegraphics[width=0.35\columnwidth]{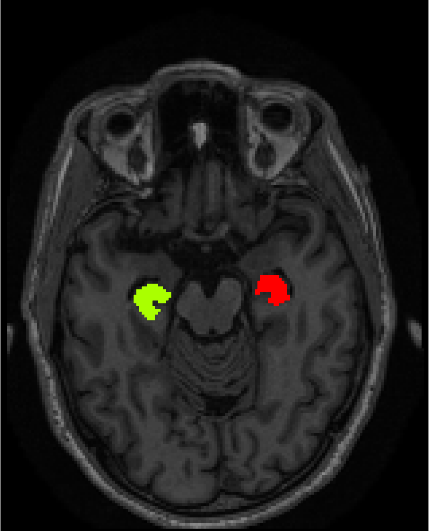}} & \includegraphics[width=0.35\columnwidth]{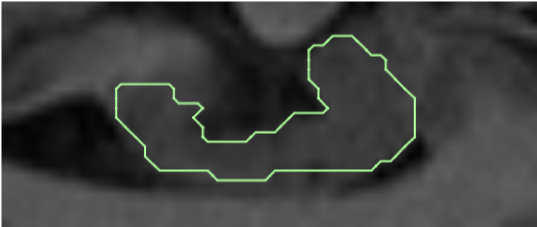} \\   
                                                                                                                                                                   &  \includegraphics[width=0.35\columnwidth]{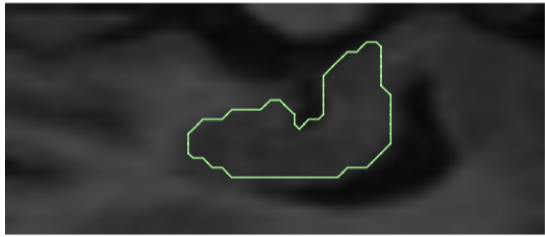}
\end{tabular}
\end{center}
\caption{\small Left: Hippocampi in a brain MR
    scan. Right: zoom-in view of two cropped slices, superimposed with
    contours of the ground-truth Hippocampus masks.}
 \label{fig:hippo3views}
 \end{figure*}

The second dataset contains brain images and the corresponding
 Hippocampus masks of 108 subjects, both downloaded from ADNI website
 (\url{adni.loni.usc.edu}). The images to be segmented are T1-weighted
 whole brain Magnetic resonance (MR) images
 (Fig. \ref{fig:hippo3views}, left), and the
 ground-truth segmentations were generated through semi-manual boundary
 delineation by human raters. In order to alleviate the data imbalance
 problem, we roughly aligned all brain MRIs, and cropped them with a
 common region that encloses the right hippocampi of all subjects,
 which led to subvolumes of size 24$\times$56$\times$48. 
 Two-dimensional slices, sized
 24$\times$56, along the axial view of all the three-dimensional
 volumes are taken as inputs in our experiments
 (Fig. \ref{fig:hippo3views}, right). The resulted
 binary segmentations of each subject are then stacked to constitute
 the final 3D segmentations for evaluation. Ten-fold cross-validation
 is used to train and evaluate the performance of the competing
 models.
 
As partially observable from
   Fig. \ref{fig:hippo3views}, human Hippocampi have rather
   complicated 3D shapes, surrounded by tissues of similar intensity
   profiles. The image contrasts in many boundary areas are very vague
   or even nonexistent, which makes accurate and reproducible
   delineations very difficult, even for human experts. In addition,
   the Hippocampus masks in ADNI were generated with a 3D surface
   fitting solution on 42 salient surface points identified by human
   raters. This fitting procedure inevitably brings noise to the
   ground-truth, imposing extra difficulty for the segmentation
   task. \citep{tong2013segmentation} (based on sparse coding \&
   dictionary learning), \citep{song2015progressive,
     wu2015hierarchical} (both based on sparse coding) and
   \citep{chen2017multiview} (based on multiview U-Net) are some of the
   recently published Hippocampus segmentation works that also used
   ADNI data. It should be noted that direct comparisons of the
   solutions are often not feasible, as different datasets, subjects
   and ground-truth setups have been used in the studies.

 \paragraph{Evaluation metrics} Totally four metrics are adopted for
 performance evaluations. \textit{Pixel accuracy} (PA), \textit{mean
   class accuracy} (MCA) and \textit{Dice ratio} are calculated in our
 experiments to evaluate the segmentation accuracies in a general
 sense. In the occlusion tests, we use an additional metric, number
 of\textit{ added holes} (AH), to assess the segmentation consistency
 of each model, in terms of topology preservation. An added hole is a
 connected, wrongly labeled area that completely resides within a
 correctly labeled segment. The AH (or more accurately, the absence of
 AH) provides a discrete metric to measure the segmentation
 consistency within a whole object.

\paragraph{Baselines}

Modified U-Nets \citep{ronneberger2015unet} are used as the baseline
models for comparison. In the MNIST experiment, the U-Net was altered
to have 5 convolution layers along the encoder path and 5
deconvolution layers along the decoder path. Pooling layers are
discarded, and we replace them with strides to reduce the feature-map
dimensions along the encoding path.  In the Hippocampus experiment,
padding is added to the original U-Net to maintain the spatial
dimension of the output of each convolution/deconvolution layer (Table
\ref{tab:baseline} in Appendix A). We only keep one convolutional
layers prior to each pooling layer. Explicit data augmentation step in
U-Net is replaced with dropout \citep{srivastava2014dropout} to reduce
overfitting. To explore the effect of receptive fields sizes, we
implement four modified U-Nets with different pooling schemes to
assign the bottom layers with various scales of feature maps. In
reporting the performance of different versions, we add a digit at the
end of U-Net to indicate the dimension of the smallest feature map
(e.g., U-Net-6 means the dimension of the feature map at the coarsest
layer is 6$\times$6).

\paragraph{Implementation details}
For the 10-digit experiment, we take a regular convolution layer as
the feature extraction layer for our TraceCaps. We feed 24 types of
capsules associated with the following primary capsule layer and each
capsule outputs a 8D vector. The primary capsule layer is followed by
the class capsule layer. The traceback pipeline is
applied between the class capsule layer and primary capsule layer. We
trained three versions of TraceCaps, with different sizes of label
maps (or dimensions of the positions) maintained in the primary
capsule layer. More specially, we use 7$\times$7, 9$\times$9 and
11$\times$11 positions, and the corresponding models are named as
TraceCaps-7, TraceCaps-9 and TraceCaps-11.

In the Hippocampus experiment, the feature extraction
  layers are adopted from first layers of the encoder in the baseline
  U-Net models. For the capsule \& traceback module, traceback-depth 1
  and 2 are implemented.  In the models with traceback depth 1, there
  are 32 capsule types in the primary capsule layers, outputting 8D
  vectors. The class capsule layer 
  in these models has one 16D vector for the 2 class capsules
  (Hippocampus and background). The traceback pipeline is between the
  class capsule layer and primary capsule layer. In the model with
  traceback-depth 2, one convolution capsule layer is inserted between
  the primary capsule layer and the class capsule layer. For this
  model, we set 64 capsules types in the primary layers with 3D
  vectors, 24 capsule types in the convolution layer with 8D vectors,
  and followed by the class capsule layer. The traceback pipeline in
  this model is along these three capsule layers.

In TraceCaps, the ratio of the two hyper-parameters
  ($\frac{\lambda_{1}}{\lambda_{2}}$) in the loss function need to be
  tuned. In the MNIST experiment, the ratio is selected from the range
  of $\{1\sim5\}$ with an interval of $1$. In the Hippocampus experiment, we
  observed that the recognition task was far more difficult, largely
  due to the greater complexity of the dataset. Accordingly, we
  selected the ratio from a larger range, $\{1\sim20\}$ with an interval
  of $1$. A large weight on recognition term is presumed to force the
  network to build more accurate part-whole relationships, which lays
  a solid foundation to produce more accurate overall segmentations.

The dynamic routing algorithm in
  \citep{hinton2017nips} is adopted in all experiments to estimate
  coupling coefficients. All networks are implemented in TensorFlow
\citep{abadi2016tensorflow} and trained with the Adam optimizer
\citep{kingma2014adam}. All experiments are run on a single Nvidia GTX
TITIAN X GPU with 12GB memory.

\subsection{Results on modified MNIST}

In this experiment, we tried different weight ratios between margin
loss and cross-entropy to evaluate the importance of the individual
loss components. We also explored setups of U-Nets with different
receptive scales.  The results are summarized in Table
\ref{tab:multidigit}. The number of parameters for
  each model is listed.  Among the baseline U-Nets, U-Net-6 (6 stands
for the dimension of the smallest feature map at the bottom layer)
gets the best results, so we only include it in the table. From the
table, it is evident that our TraceCaps outperform the best U-Net
model in all metrics for this dataset. We can also tell that two
factors, dimension of primary capsule layer (7/9/11) and weights for
loss terms, affect the performance of our TraceCaps.

\begin{center}
\scalebox{0.75}{
  \begin{tabu}{l c c | c  c c }
    \hline
    \hline
    Method & \makecell{Loss Weights \\ ($\lambda_{1}$, $\lambda_{2}$)} & \makecell{Number of \\ Training Parameters} & PA  &  Mean Accuracy & Dice Ratio\\
    
    \hline
    TraceCaps-7 & 1, 1 & 5.09M &\textbf{99.06} & \textbf{99.07}  & $99.19  \pm 5.81$\\
    TraceCaps-9  & 2, 1  & 3.40M &99.01  & 99.02  & $99.23 \pm 5.22$ \\
    TraceCaps-9 & 1, 1 &3.40M & 99.04  & 99.05  & \textbf{99.27 $\pm$ 5.05}\\
    TraceCaps-11 & 1, 1 & 2.24M &98.31  & 98.31  & $98.89 \pm 5.34$\\
    \hline\hline
    U-Net-6     & - & 3.14M  &98.04  & 98.03  & $95.63 \pm 6.13$\\
    \hline
  \end{tabu}}
  \captionof{table}{Results on the 10-digit experiment. Refer to text for details.}
\label{tab:multidigit}
\end{center}

\subsection{Results on Hippocampus dataset}
\paragraph{Effects of size of the primary capsule layer} In order to
explore how the dimension of primary layer (number of positions) plays
a role in the segmentation procedure, we train different versions of
TraceCaps, as we do for MNIST experiments. The results on one
particular split of the cross-validation test is shown in Table
\ref{tab:hippoeffect}. In column 3, we list the
  traceback-size and traceback-depth of each model. From the results,
it appears that the size of primary layer size and the weights of the
loss terms both affect the results to certain extent.

\begin{center}
\scalebox{0.85}{
\begin{tabu}{l c  c  c| c }
    \hline
    \hline
    Method & \makecell{Number of \\ Training Parameters} & \makecell{ Traceback-size \\ and Depth}  & \makecell{Loss Weights \\ ($\lambda_{1}$, $\lambda_{2}$)} & Dice Ratio \\
    
    \hline
    TraceCaps &1.50M &2$\times$6, 1 & (7, 1)  &$87.51 \pm 2.378$ \\
    TraceCaps & 1.57M &4$\times$12, 1 & (10, 1) & $ 88.05\pm 2.541$ \\
    TraceCaps &1.90M &4$\times$12, 2 & (10, 1) & $ 88.19\pm 1.874$ \\
    TraceCaps & 3.02M &4$\times$20, 1 &  (15, 1) &\textbf{88.86 $\pm$ 1.628} \\
    \hline
\end{tabu}}
\captionof{table}{Effect of the traceback pipeline on
  performance of TraceCaps models. Note that the results are reported
  based on one particular split of the dataset. Refer to text for
  details.}
\label{tab:hippoeffect}
\end{center}

For our TraceCaps, when the primary capsule size set to 4$\times$20
and loss weights set to 15:1, we obtain the most accurate segmentation
results. Reducing the dimension of primary layer, as well as
decreasing the contribution of margin-loss, seems worsening the model
performance. One possible explanation is that when the the dimension
of primary layer is set to a very small number, the positions of the
primary capsules may not be precise enough to represent the visual
entities at the input image level.

For the contribution of margin loss increasing it should make the
network strive to obtain more accurate overall recognition. With our
mathematically rigorous traceback, higher recognition accuracy would
translate into more precise membership labelling at individual pixels.

For U-Nets, with a fixed input size, the feature map size of the
bottom layer is the reciprocal of the size of the receptive field of
this layer.  Setting the feature map size to a small number would
allow the information of pixels from a broader range to be integrated,
leading to improved class decision and boundary localization.
We test the U-Nets using the same data as in Table
  \ref{tab:hippoeffect}, and this trend can be observed in Table
\ref{tab:uneteffect} from U-Net-3$\times$7 to U-Net-4$\times$20.

\begin{center}
\begin{tabu}{l c  c  c }
    \hline
    \hline
    Method & \makecell{Number of \\ Training Parameters} & \makecell{ Smallest \\ Feature Map}  & Dice Ratio \\
    \hline
    U-Net & 1.14M &3$\times$7 &\textbf{88.41 $\pm$ 1.707} \\
    U-Net & 0.95M &4$\times$12 &   $88.27 \pm 1.778$ \\
    U-Net & 2.19M &4$\times$20 &  $87.68 \pm 2.219$  \\
    \hline
\end{tabu}
\captionof{table}{Effect of the smallest feature map on performance of
  baseline U-Net models. Refer to text for details.}
\label{tab:uneteffect}
\end{center}

\paragraph{Overall average performance} Through the one-split
validation, we identified the potentially best setups for both
TraceCaps and modified U-Net, which are TraceCaps- 4$\times$20 and
U-Net-3$\times$7, respectively (refer to Table \ref{tab:layerconfig}
in Appendix A for detailed layer configurations). We then carried out
a nine-fold cross-validation on the entire 108 data points.  Table
\ref{tab:hippo} shows the segmentation accuracies of TraceCaps and
U-Net in all the nine folds, followed by the averages. As evident,
TraceCaps obtain a higher Dice ratio in each fold.  On average, the
Dice ratio for TraceCaps- 4$\times$20 is 87.25 with a standard
deviation 5.05, while U-Net-3$\times$7 obtaines 86.23 with a standard
deviation 2.19. 
model outperforms the best U-Net model.

\begin{center}
\begin{tabu}{c c c  }
    \hline
    \hline
    Fold & TraceCaps & U-Net  \\    
    \hline
    1 &   \makecell{ \textbf{88.86} $\pm$ 1.628}  &   \makecell{88.41 $\pm$ 1.707 }  \\
    2 &   \makecell{ \textbf{87.84} $\pm$ 5.674} &  \makecell{86.03 $\pm$ 3.347 } \\
    3 &   \makecell{ \textbf{86.53} $\pm$4.678} & \makecell{ 85.42 $\pm$ 3.005 } \\
    4 &   \makecell{\textbf{85.73}$\pm$ 7.352}  & \makecell{85.21 $\pm$ 2.301 } \\
    5 &   \makecell{\textbf{87.09}$\pm$ 4.190} &  \makecell{86.53 $\pm$ 2.159 }  \\
    6 &   \makecell{\textbf{83.56} $\pm$ 1.257} &  \makecell{82.34$\pm$ 1.369}  \\
    7 &   \makecell{\textbf{88.57} $\pm$ 1.783} &  \makecell{ 87.14 $\pm$ 1.374 } \\
    8 &    \makecell{\textbf{88.23}$\pm$ 4.506} &  \makecell{86.96 $\pm$ 2.071 } \\
    9 &   \makecell{\textbf{88.82}$\pm$ 3.005} & \makecell{88.03 $\pm$ 2.333 }  \\
    \hline
    Average & \makecell{\textbf{87.25} $\pm$ 3.786} & \makecell{86.23 $\pm$ 2.193 } \\
    \hline
\end{tabu}
\captionof{table}{Hippocampus segmentation accuracies on 9-fold
    cross-validaton.}
\label{tab:hippo}
\end{center}

\subsection{Occlusion test}

One of the fundamental differences between our TraceCaps model and
FCNs lies in the fact that TraceCaps does segmentation and
recognition simultaneously. The recognition subtask, if conducted well
in practice, would equip TraceCaps with an additional power in
dealing with adversarial situations, which include input images with
occlusions. In order to evaluate and demonstrate the effect, we design
an occlusion test as follows.

We train our model on images of two easily confused digits (e.g. 0 and
8) with the modified MNIST samples. For the test samples, however, we
generate occlusions in each image by setting the intensities of
several horizontal lines around the image centers to black. The
segmentation results are shown in Table \ref{tab:occlu1}. It is
evident that TraceCaps achieve significantly higher accuracies than
the best baseline model.

\begin{center}
\scalebox{0.85}{
 \begin{tabular}{ l c c c c | c  c | c  c }
    \hline
    \hline
    \multirow{2}{*}{Method} & \multirow{2}{*}{\makecell{Loss Weights \\ ($\lambda_{1}$, $\lambda_{2}$)}}  &\multirow{2}{*}{PA} &\multirow{2}{*}{Dice} & \multirow{2}{*}{AH}  & \multicolumn{2}{c|}{0} & \multicolumn{2}{c}{8} \\  
    \cline{6-9} &&&&& Dice & AH    & Dice & AH \\
    
    \hline
    TraceCaps-9 & (2, 1) &   \textbf{93.49} & \textbf{95.44} & \textbf{264} &93.89  &  185 & \textbf{96.99} & \textbf{79} \\
    TraceCaps-9 & (3, 1) &  92.73 & 95.05 & 270 & \textbf{95.36} & \textbf{123}  & 94.74 &  147\\
    U-Net-6 & - &91.17 & 91.20 & 300 & 92.59 & 123  & 89.82 &  177  \\
    \hline
  \end{tabular}}
  \captionof{table}{Results on occluded test. Refer to text for details.}
  \label{tab:occlu1}
\end{center}

Table \ref{tab:occlu2} shows several representative prediction images
from the occlusion test. Pixels classified into digit 8 are shown in
green color and red color pixels have been classified as digit 0. In
the first column, U-Net-6 generates rather precise boundary
localization but makes totally wrong labellings for all pixels. Our
TraceCaps gets both aspects, overall classification and boundary
localization, very well worked out. In column 2, U-Net-6 makes the
same mistake for the entire digit, while TraceCaps put correct labels
for vast majority of the pixels. U-Net-6 performs correctly for the
input in column 3. Overall, TraceCaps appears to be able to generate
more robust and consistent segmentation results, and this capability
should be attributed to the recognition component and the part-whole
mechanism built in our model.

\begin{center}
  \begin{tabular}{ l  c c  c  c c  c}
    \hline
    \hline
         \parbox[c]{1em}{    } &  \parbox[c]{1em}{8} & \parbox[c]{1em}{8} & \parbox[c]{2em}{0} \\ \hline
    Ground-truth & \parbox[c]{5em}{\includegraphics[width=0.125\columnwidth]{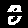} } & \parbox[c]{5em}{\includegraphics[width=0.125\columnwidth]{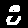} }  & \parbox[c]{5em}{\includegraphics[width=0.125\columnwidth]{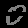} }  \\
    TraceCaps-9 &    \parbox[c]{5em}{\includegraphics[width=0.125\columnwidth]{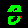} } & \parbox[c]{5em}{\includegraphics[width=0.125\columnwidth]{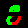} }  & \parbox[c]{5em}{\includegraphics[width=0.125\columnwidth]{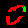} }  \\ 
    U-Net-6 & \parbox[c]{5em}{\includegraphics[width=0.125\columnwidth]{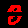} } & \parbox[c]{5em}{\includegraphics[width=0.125\columnwidth]{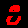} }  & \parbox[c]{5em}{\includegraphics[width=0.125\columnwidth]{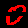} }   \\
    \hline
   
  \end{tabular}
  \captionof{table}{Prediction results from the occlusion test. Refer to text for details.}
\label{tab:occlu2}
\end{center}

\section{Discussion and Related Work}
Since the inception of the first FCN model, FCN variants have become
the most popular solutions for semantic segmentation. Compared with
traditional non-deep learning techniques as well as patch-based CNN
solutions, the efficacy of FCNs should be attributed, in large part,
to their capability to process information from various spatial
scales.

\paragraph{FCN models} While powerful, FCNs also have some inherent
limitations. Originated from CNNs, FCNs tend to lose precise spatial
information along the pooling operations/layers, and inconsistent
pixel labelings can be resulted from the limited receptive field at
each neuron. A number of approaches have explored to add the
contextual information back to the CNN component. Conditional Random
Fields (CRFs) based solutions \citep{chen2014crf,chen2018deeplab}
refine the final segmentation results with a post-processing stage to
boost its ability to capture fine-grained details. Dilated
convolutions \citep{yu2015dilate,yu2017dilated} are utilized to take
advantage of the fact that the receptive fields of systematic dilation
can be expanded exponentially without losing resolution. Fusion of
global features extracted at lower layers and local features from
higher layers through skip connections have also been well studied
\citep{long2015fully,ronneberger2015unet}. To produce segmentation
results with certain shape constraints,
\citep{ravishankar2017shape,chen2013shape} integrate shape priors into
existing deep learning framework with shape-based loss terms. New
models have emerged in the past {\color{black} two years} or so, such as
multi-resolution fusion by RefineNet \citep{lin2017refinenet}, pyramid
pooling module in PSPNet \citep{zhao2017pyramid}, Clusternet
\citep{lalonde2018clusternet}, atrous convolution and fully connected
CRFs \citep{zhang2018exfuse}, Y-Net \citep{mehta2018net},
{\color{black}HRNet \citep{sun2019high}, Panoptic-deeplab
  \citep{cheng2020panoptic}, and models trained with self-training
  \citep{zoph2020rethinking}}.


\paragraph{Capsules and capsule nets}
The history of the notion of capsules can be dated back to
\citep{hinton1981capsule}, in which the author proposed to describe an
object over the viewpoint-invariant spatial relationships between the
object and its parts. This model should be regarded as the first
theoretical prototype of the recent capsule nets
{\color{black}\citep{hinton2017nips,hinton2018ICLR,hinton2019Nips}}. \cite{zemel1990traffic}
added probability components into this prototype and encoded both
probabilities and viewpoint-invariance in a fully-connected neural
network. The instantiation parameters of Zemel's model were
hand-crafted from input data.

The issue of how to initialize instantiation parameters was partially
addressed in \citep{hinton2011autoencoder}, but their model requires
transformation matrices to be input externally. Regarding the routing
algorithms, \cite{zemel1990traffic} introduced the notion that an
object can be activated by combining the predictions from its parts,
which was materialized as an averaging operation in
\citep{hinton2011autoencoder}.  \cite{hinton2017nips,hinton2018ICLR}
took a step forward and resorted to routing-by-agreement algorithms to
simultaneously activate objects and set up part-whole relationships.

Recently, more routing schemes have been proposed, including new
mechanisms based on attention
\citep{choi2019attention,duarte2019capsulevos}, distance measures
\citep{lenssen2018group} and weighted kernel density estimation
\citep{zhang2018fast}. In \citep{mcIntosh2020vt} and
\citep{lalonde2019encoding}, visual-text coupling has also been
explored to design multi-modality routing mechanisms for video
segmentation and lung cancer predictions, respectively.

On top of supervised capsule nets, \cite{hinton2019Nips} introduced
unsupervised capsule nets in 2019, which adopted the approachs of
capsule initiation in \citep{hinton2011autoencoder} and routing
mechanism in \citep{hinton2018ICLR}, but relies on set transformer
(Lee et al., 2019) to infer whole-part relationships.  Application
wise, capsule nets have been explored in tasks including text
classification \citep{zhao2018investigating}, action detection
\citep{duarte2018video}, visual question answering
\citep{zhou2019dynamic} and medical image analysis
\citep{lalonde2020diagnosing,afshar2018brain,iesmantas2018convolutional,mobiny2019automated,pal2018capsdemm},
among others.


\paragraph{Capsule-based segmentation solutions}

{\color{black} \citep{lalonde2018capsules} and
  \citep{bonheur2019matwo} are two capsule-based models proposed for
  the segmentation task. In their network called {\it SegCaps},
  \cite{lalonde2018capsules} made two modifications to the original
  dynamic routing algorithm}, such that the number of parameters is
greatly reduced, which enables the model to operate on large images
sizes. Deconvolutional capsule layers were devised and appended
directly to convolutional capsule layers to restore the original
resolution. Similar to FCNs, this model utilizes skip connections to
integrate contextual information from different spatial
scales. {\color{black}\cite{bonheur2019matwo} introduced
  \textit{Matwo-CapsNet}, which is a U-shaped (similar to U-Net)
  network enabled by a new type of capsules that encode pose and
  appearance information and a {\it dual routing} scheme to integrate
  such information}.

Our TraceCaps, however, is designed under a different paradigm. Taking
the extractable part-whole information as the foundation, label
assignments in TraceCaps are explicit, interpretable, and
mathematically well-grounded. The decoder of our model is much simpler
than that of \citep{lalonde2018capsules} model, with no trainable
parameters (up to deconvolution layers) and no capsule-wise routing
needed. Because of the analytical derivations, our model has great
potential to be applied to many other applications, including object
localization, detection and visualization of heap maps.
{\color{black} While Matwo-CapsNet \citep{bonheur2019matwo} was
  designed to improve SegCaps, the performance of both models was
  reported to be mostly on par with the state-of-the-art solutions
  (i.e. different variants of the U-Net).  As a contrast, our
  TraceCaps significantly outperforms U-Net in a rather consistent
  fashion throughout our experiments. Please note that we do not
  intend to make a head-to-head comparison among these models, as the
  studies were conducted and reported based on different datasets and
  setups. Nevertheless, the consistent performance improvements made
  by our model over U-Net should be considered as a strong evidence of
  its technical merits.}


\paragraph{From capsule-based recognition to capsule-based
  segmentation and beyond}
Capsule nets have an assumption in their constructions:
\textquotedblleft at each location in the image, there is at most one
instance of the type of entity that a capsule
represents.\textquotedblright \citep{hinton2017nips}.  Essentially,
capsule nets, including the internal part-whole relationships, are
built on concrete, physical object instances. With this assumption,
capsule nets perform well when there is at most one instance of a
category in the image, such as on MNIST and smallNORB
\citep{lecun2004learning}. For the image data that have multiple
instances of same classes, capsule nets have no guarantee to
outperform CNNs in recognition accuracy.

The traceback pipeline in our TraceCaps does not rely on the this
one-instance assumption. However, the current version TraceCaps is
designed based on the capsule nets, where the margin loss is specified
with outputs of the class capsule layer. The inability of capsule nets
in handling multi-instance-same-class cases also limits the capacity
of our segmentation model, for certain type of dataset. Regarding the
remedies, we believe efforts can be push forward along three
fronts. The first one would rely on the continued development of
capsule and capsule nets to overcome their limitations. The second
possibility would be designing traceback solutions that can circumvent
the multi-instance constraint. For example, if the class capsule layer
can somehow be removed from the traceback pipeline, potentially we
could accommodate multiple instances of the same category, starting at
a convolutional capsule layer. The third group of remedies could be
developed along pre-processing direction. For example, object
detection can be carried out first, followed by TraceCaps to generate
accuracy segmentations within individual regions.

\bigskip


\section{Acknowledgments}

This study was financially supported by grants by Stocker Endowment,
Charles River Labortories, Moores Alzheimer Research Endowment,
Sanders-Brown Center on Aging and University of Kentucky College of
Medicine.


Data collection and sharing for this project was funded by the
Alzheimer's Disease Neuroimaging Initiative (ADNI) (National
Institutes of Health Grant U01 AG024904) and DOD ADNI (Department of
Defense award number W81XWH-12-2-0012). ADNI is funded by the National
Institute on Aging, the National Institute of Biomedical Imaging and
Bioengineering, and through generous contributions from the following:
AbbVie, Alzheimer’s Association; Alzheimer’s Drug Discovery
Foundation; Araclon Biotech; BioClinica, Inc.; Biogen; Bristol-Myers
Squibb Company; CereSpir, Inc.; Eisai Inc.; Elan Pharmaceuticals,
Inc.; Eli Lilly and Company; EuroImmun; F. Hoffmann-La Roche Ltd and
its affiliated company Genentech, Inc.; Fujirebio; GE Healthcare;
IXICO Ltd.; Janssen Alzheimer Immunotherapy Research $\&$ Development,
LLC.; Johnson $\&$ Johnson Pharmaceutical Research $\&$ Development
LLC.; Lumosity; Lundbeck; Merck $\&$ Co., Inc.; Meso Scale
Diagnostics, LLC.; NeuroRx Research; Neurotrack Technologies; Novartis
Pharmaceuticals Corporation; Pfizer Inc.; Piramal Imaging; Servier;
Takeda Pharmaceutical Company; and Transition Therapeutics. The
Canadian Institutes of Health Research is providing funds to support
ADNI clinical sites in Canada. Private sector contributions are
facilitated by the Foundation for the National Institutes of Health
(www.fnih.org). The grantee organization is the Northern California
Institute for Research and Education, and the study is coordinated by
the Alzheimer's Disease Cooperative Study at the University of
California, San Diego.  ADNI data are disseminated by the Laboratory
for Neuro Imaging at the University of Southern California.





\bibliography{elsa_reference}

\begin{thebibliography}{54}
\expandafter\ifx\csname natexlab\endcsname\relax\def\natexlab#1{#1}\fi
\providecommand{\url}[1]{\texttt{#1}}
\providecommand{\href}[2]{#2}
\providecommand{\path}[1]{#1}
\providecommand{\DOIprefix}{doi:}
\providecommand{\ArXivprefix}{arXiv:}
\providecommand{\URLprefix}{URL: }
\providecommand{\Pubmedprefix}{pmid:}
\providecommand{\doi}[1]{\href{http://dx.doi.org/#1}{\path{#1}}}
\providecommand{\Pubmed}[1]{\href{pmid:#1}{\path{#1}}}
\providecommand{\bibinfo}[2]{#2}
\ifx\xfnm\relax \def\xfnm[#1]{\unskip,\space#1}\fi
\bibitem[{Abadi et~al.(2016)Abadi, Agarwal, Barham, Brevdo, Chen, Citro,
  Corrado, Davis, Dean, Devin et~al.}]{abadi2016tensorflow}
\bibinfo{author}{Abadi, M.}, \bibinfo{author}{Agarwal, A.},
  \bibinfo{author}{Barham, P.}, \bibinfo{author}{Brevdo, E.},
  \bibinfo{author}{Chen, Z.}, \bibinfo{author}{Citro, C.},
  \bibinfo{author}{Corrado, G.S.}, \bibinfo{author}{Davis, A.},
  \bibinfo{author}{Dean, J.}, \bibinfo{author}{Devin, M.}, et~al.,
  \bibinfo{year}{2016}.
\newblock \bibinfo{title}{Tensorflow: Large-scale machine learning on
  heterogeneous distributed systems}.
\newblock \bibinfo{journal}{arXiv preprint arXiv:1603.04467} .
\bibitem[{Afshar et~al.(2018)Afshar, Mohammadi and
  Plataniotis}]{afshar2018brain}
\bibinfo{author}{Afshar, P.}, \bibinfo{author}{Mohammadi, A.},
  \bibinfo{author}{Plataniotis, K.N.}, \bibinfo{year}{2018}.
\newblock \bibinfo{title}{Brain tumor type classification via capsule
  networks}, in: \bibinfo{booktitle}{2018 25th IEEE International Conference on
  Image Processing (ICIP)}, \bibinfo{organization}{IEEE}. pp.
  \bibinfo{pages}{3129--3133}.
\bibitem[{Badrinarayanan et~al.(2017)Badrinarayanan, Kendall and
  Cipolla}]{segnet}
\bibinfo{author}{Badrinarayanan, V.}, \bibinfo{author}{Kendall, A.},
  \bibinfo{author}{Cipolla, R.}, \bibinfo{year}{2017}.
\newblock \bibinfo{title}{Segnet: A deep convolutional encoder-decoder
  architecture for image segmentation}.
\newblock \bibinfo{journal}{IEEE Transactions on Pattern Analysis and Machine
  Intelligence} .
\bibitem[{Bonheur et~al.(2019)Bonheur, {\v{S}}tern, Payer, Pienn, Olschewski
  and Urschler}]{bonheur2019matwo}
\bibinfo{author}{Bonheur, S.}, \bibinfo{author}{{\v{S}}tern, D.},
  \bibinfo{author}{Payer, C.}, \bibinfo{author}{Pienn, M.},
  \bibinfo{author}{Olschewski, H.}, \bibinfo{author}{Urschler, M.},
  \bibinfo{year}{2019}.
\newblock \bibinfo{title}{Matwo-capsnet: a multi-label semantic segmentation
  capsules network}, in: \bibinfo{booktitle}{International Conference on
  Medical Image Computing and Computer-Assisted Intervention},
  \bibinfo{organization}{Springer}. pp. \bibinfo{pages}{664--672}.
\bibitem[{Chen et~al.(2013)Chen, Yu, Hu and Zeng}]{chen2013shape}
\bibinfo{author}{Chen, F.}, \bibinfo{author}{Yu, H.}, \bibinfo{author}{Hu, R.},
  \bibinfo{author}{Zeng, X.}, \bibinfo{year}{2013}.
\newblock \bibinfo{title}{Deep learning shape priors for object segmentation},
  in: \bibinfo{booktitle}{Computer Vision and Pattern Recognition (CVPR), 2013
  IEEE Conference on}, \bibinfo{organization}{IEEE}. pp.
  \bibinfo{pages}{1870--1877}.
\bibitem[{Chen et~al.(2014)Chen, Papandreou, Kokkinos, Murphy and
  Yuille}]{chen2014crf}
\bibinfo{author}{Chen, L.C.}, \bibinfo{author}{Papandreou, G.},
  \bibinfo{author}{Kokkinos, I.}, \bibinfo{author}{Murphy, K.},
  \bibinfo{author}{Yuille, A.L.}, \bibinfo{year}{2014}.
\newblock \bibinfo{title}{Semantic image segmentation with deep convolutional
  nets and fully connected crfs}.
\newblock \bibinfo{journal}{arXiv preprint arXiv:1412.7062} .
\bibitem[{Chen et~al.(2018)Chen, Papandreou, Kokkinos, Murphy and
  Yuille}]{chen2018deeplab}
\bibinfo{author}{Chen, L.C.}, \bibinfo{author}{Papandreou, G.},
  \bibinfo{author}{Kokkinos, I.}, \bibinfo{author}{Murphy, K.},
  \bibinfo{author}{Yuille, A.L.}, \bibinfo{year}{2018}.
\newblock \bibinfo{title}{Deeplab: Semantic image segmentation with deep
  convolutional nets, atrous convolution, and fully connected crfs}.
\newblock \bibinfo{journal}{IEEE transactions on pattern analysis and machine
  intelligence} \bibinfo{volume}{40}, \bibinfo{pages}{834--848}.
\bibitem[{Chen et~al.(2017)Chen, Shi, Wang, Zhang, Smith and
  Liu}]{chen2017multiview}
\bibinfo{author}{Chen, Y.}, \bibinfo{author}{Shi, B.}, \bibinfo{author}{Wang,
  Z.}, \bibinfo{author}{Zhang, P.}, \bibinfo{author}{Smith, C.D.},
  \bibinfo{author}{Liu, J.}, \bibinfo{year}{2017}.
\newblock \bibinfo{title}{Hippocampus segmentation through multi-view ensemble
  convnets}, in: \bibinfo{booktitle}{Biomedical Imaging (ISBI 2017), 2017 IEEE
  14th International Symposium on}, \bibinfo{organization}{IEEE}. pp.
  \bibinfo{pages}{192--196}.
\bibitem[{Cheng et~al.(2020)Cheng, Collins, Zhu, Liu, Huang, Adam and
  Chen}]{cheng2020panoptic}
\bibinfo{author}{Cheng, B.}, \bibinfo{author}{Collins, M.D.},
  \bibinfo{author}{Zhu, Y.}, \bibinfo{author}{Liu, T.}, \bibinfo{author}{Huang,
  T.S.}, \bibinfo{author}{Adam, H.}, \bibinfo{author}{Chen, L.C.},
  \bibinfo{year}{2020}.
\newblock \bibinfo{title}{Panoptic-deeplab: A simple, strong, and fast baseline
  for bottom-up panoptic segmentation}, in: \bibinfo{booktitle}{Proceedings of
  the IEEE/CVF Conference on Computer Vision and Pattern Recognition}, pp.
  \bibinfo{pages}{12475--12485}.
\bibitem[{Choi et~al.(2019)Choi, Seo, Im and Kang}]{choi2019attention}
\bibinfo{author}{Choi, J.}, \bibinfo{author}{Seo, H.}, \bibinfo{author}{Im,
  S.}, \bibinfo{author}{Kang, M.}, \bibinfo{year}{2019}.
\newblock \bibinfo{title}{Attention routing between capsules}, in:
  \bibinfo{booktitle}{Proceedings of the IEEE International Conference on
  Computer Vision Workshops}, pp. \bibinfo{pages}{0--0}.
\bibitem[{Cohen and Welling(2016)}]{cohen2016group}
\bibinfo{author}{Cohen, T.}, \bibinfo{author}{Welling, M.},
  \bibinfo{year}{2016}.
\newblock \bibinfo{title}{Group equivariant convolutional networks}, in:
  \bibinfo{booktitle}{International conference on machine learning}, pp.
  \bibinfo{pages}{2990--2999}.
\bibitem[{Cohen et~al.(2018)Cohen, Geiger, K{\"o}hler and
  Welling}]{cohen2018spherical}
\bibinfo{author}{Cohen, T.S.}, \bibinfo{author}{Geiger, M.},
  \bibinfo{author}{K{\"o}hler, J.}, \bibinfo{author}{Welling, M.},
  \bibinfo{year}{2018}.
\newblock \bibinfo{title}{Spherical cnns}.
\newblock \bibinfo{journal}{arXiv preprint arXiv:1801.10130} .
\bibitem[{Duarte et~al.(2018)Duarte, Rawat and Shah}]{duarte2018video}
\bibinfo{author}{Duarte, K.}, \bibinfo{author}{Rawat, Y.S.},
  \bibinfo{author}{Shah, M.}, \bibinfo{year}{2018}.
\newblock \bibinfo{title}{Videocapsulenet: A simplified network for action
  detection}, in: \bibinfo{booktitle}{Advances in neural information processing
  systems}, pp. \bibinfo{pages}{7610--7619}.
\bibitem[{Duarte et~al.(2019)Duarte, Rawat and Shah}]{duarte2019capsulevos}
\bibinfo{author}{Duarte, K.}, \bibinfo{author}{Rawat, Y.S.},
  \bibinfo{author}{Shah, M.}, \bibinfo{year}{2019}.
\newblock \bibinfo{title}{Capsulevos: Semi-supervised video object segmentation
  using capsule routing}, in: \bibinfo{booktitle}{Proceedings of the IEEE
  International Conference on Computer Vision}, pp.
  \bibinfo{pages}{8480--8489}.
\bibitem[{Hinton et~al.(2000)Hinton, Ghahramani and Teh}]{hinton2000learning}
\bibinfo{author}{Hinton, G.E.}, \bibinfo{author}{Ghahramani, Z.},
  \bibinfo{author}{Teh, Y.W.}, \bibinfo{year}{2000}.
\newblock \bibinfo{title}{Learning to parse images}, in:
  \bibinfo{booktitle}{Advances in neural information processing systems}, pp.
  \bibinfo{pages}{463--469}.
\bibitem[{Hinton et~al.(2011)Hinton, Krizhevsky and
  Wang}]{hinton2011autoencoder}
\bibinfo{author}{Hinton, G.E.}, \bibinfo{author}{Krizhevsky, A.},
  \bibinfo{author}{Wang, S.D.}, \bibinfo{year}{2011}.
\newblock \bibinfo{title}{Transforming auto-encoders}, in:
  \bibinfo{booktitle}{International Conference on Artificial Neural Networks},
  \bibinfo{organization}{Springer}. pp. \bibinfo{pages}{44--51}.
\bibitem[{Hinton et~al.(2018)Hinton, Sabour and Frosst}]{hinton2018ICLR}
\bibinfo{author}{Hinton, G.E.}, \bibinfo{author}{Sabour, S.},
  \bibinfo{author}{Frosst, N.}, \bibinfo{year}{2018}.
\newblock \bibinfo{title}{Matrix capsules with {EM} routing}, in:
  \bibinfo{booktitle}{International Conference on Learning Representations}.
\bibitem[{Hinton(1981)}]{hinton1981capsule}
\bibinfo{author}{Hinton, G.F.}, \bibinfo{year}{1981}.
\newblock \bibinfo{title}{A parallel computation that assigns canonical
  object-based frames of reference}, in: \bibinfo{booktitle}{Proceedings of the
  7th international joint conference on Artificial intelligence-Volume 2},
  \bibinfo{organization}{Morgan Kaufmann Publishers Inc.}. pp.
  \bibinfo{pages}{683--685}.
\bibitem[{Iesmantas and Alzbutas(2018)}]{iesmantas2018convolutional}
\bibinfo{author}{Iesmantas, T.}, \bibinfo{author}{Alzbutas, R.},
  \bibinfo{year}{2018}.
\newblock \bibinfo{title}{Convolutional capsule network for classification of
  breast cancer histology images}, in: \bibinfo{booktitle}{International
  Conference Image Analysis and Recognition}, \bibinfo{organization}{Springer}.
  pp. \bibinfo{pages}{853--860}.
\bibitem[{Kingma and Ba(2014)}]{kingma2014adam}
\bibinfo{author}{Kingma, D.P.}, \bibinfo{author}{Ba, J.}, \bibinfo{year}{2014}.
\newblock \bibinfo{title}{Adam: A method for stochastic optimization}.
\newblock \bibinfo{journal}{arXiv preprint arXiv:1412.6980} .
\bibitem[{Kosiorek et~al.(2019)Kosiorek, Sabour, Teh and
  Hinton}]{hinton2019Nips}
\bibinfo{author}{Kosiorek, A.}, \bibinfo{author}{Sabour, S.},
  \bibinfo{author}{Teh, Y.W.}, \bibinfo{author}{Hinton, G.E.},
  \bibinfo{year}{2019}.
\newblock \bibinfo{title}{Stacked capsule autoencoders}, in:
  \bibinfo{booktitle}{Advances in Neural Information Processing Systems}, pp.
  \bibinfo{pages}{15512--15522}.
\bibitem[{Krizhevsky et~al.(2012)Krizhevsky, Sutskever and
  Hinton}]{krizhevsky2012alexnet}
\bibinfo{author}{Krizhevsky, A.}, \bibinfo{author}{Sutskever, I.},
  \bibinfo{author}{Hinton, G.E.}, \bibinfo{year}{2012}.
\newblock \bibinfo{title}{Imagenet classification with deep convolutional
  neural networks}, in: \bibinfo{booktitle}{Advances in neural information
  processing systems}, pp. \bibinfo{pages}{1097--1105}.
\bibitem[{LaLonde and Bagci(2018)}]{lalonde2018capsules}
\bibinfo{author}{LaLonde, R.}, \bibinfo{author}{Bagci, U.},
  \bibinfo{year}{2018}.
\newblock \bibinfo{title}{Capsules for object segmentation}.
\newblock \bibinfo{journal}{arXiv preprint arXiv:1804.04241} .
\bibitem[{LaLonde et~al.(2020)LaLonde, Kandel, Spampinato, Wallace and
  Bagci}]{lalonde2020diagnosing}
\bibinfo{author}{LaLonde, R.}, \bibinfo{author}{Kandel, P.},
  \bibinfo{author}{Spampinato, C.}, \bibinfo{author}{Wallace, M.B.},
  \bibinfo{author}{Bagci, U.}, \bibinfo{year}{2020}.
\newblock \bibinfo{title}{Diagnosing colorectal polyps in the wild with capsule
  networks}, in: \bibinfo{booktitle}{2020 IEEE 17th International Symposium on
  Biomedical Imaging (ISBI)}, \bibinfo{organization}{IEEE}. pp.
  \bibinfo{pages}{1086--1090}.
\bibitem[{LaLonde et~al.(2019)LaLonde, Torigian and
  Bagci}]{lalonde2019encoding}
\bibinfo{author}{LaLonde, R.}, \bibinfo{author}{Torigian, D.},
  \bibinfo{author}{Bagci, U.}, \bibinfo{year}{2019}.
\newblock \bibinfo{title}{Encoding visual attributes in capsules for
  explainable medical diagnoses}.
\newblock \bibinfo{journal}{arXiv} , \bibinfo{pages}{arXiv--1909}.
\bibitem[{LaLonde et~al.(2018)LaLonde, Zhang and Shah}]{lalonde2018clusternet}
\bibinfo{author}{LaLonde, R.}, \bibinfo{author}{Zhang, D.},
  \bibinfo{author}{Shah, M.}, \bibinfo{year}{2018}.
\newblock \bibinfo{title}{Clusternet: Detecting small objects in large scenes
  by exploiting spatio-temporal information}, in: \bibinfo{booktitle}{Computer
  Vision and Pattern Recognition}.
\bibitem[{LeCun and Cortes(1998)}]{lecun1998mnist}
\bibinfo{author}{LeCun, Y.}, \bibinfo{author}{Cortes, C.},
  \bibinfo{year}{1998}.
\newblock \bibinfo{title}{The mnist database of handwritten digits}.
\newblock \bibinfo{journal}{http://yann. lecun. com/exdb/mnist/} .
\bibitem[{LeCun et~al.(2004)LeCun, Huang and Bottou}]{lecun2004learning}
\bibinfo{author}{LeCun, Y.}, \bibinfo{author}{Huang, F.J.},
  \bibinfo{author}{Bottou, L.}, \bibinfo{year}{2004}.
\newblock \bibinfo{title}{Learning methods for generic object recognition with
  invariance to pose and lighting}, in: \bibinfo{booktitle}{Computer Vision and
  Pattern Recognition, 2004. CVPR 2004. Proceedings of the 2004 IEEE Computer
  Society Conference on}, \bibinfo{organization}{IEEE}. pp.
  \bibinfo{pages}{II--104}.
\bibitem[{Lenssen et~al.(2018)Lenssen, Fey and Libuschewski}]{lenssen2018group}
\bibinfo{author}{Lenssen, J.E.}, \bibinfo{author}{Fey, M.},
  \bibinfo{author}{Libuschewski, P.}, \bibinfo{year}{2018}.
\newblock \bibinfo{title}{Group equivariant capsule networks}, in:
  \bibinfo{booktitle}{Advances in neural information processing systems}, pp.
  \bibinfo{pages}{8844--8853}.
\bibitem[{Lin et~al.(2017)Lin, Milan, Shen and Reid}]{lin2017refinenet}
\bibinfo{author}{Lin, G.}, \bibinfo{author}{Milan, A.}, \bibinfo{author}{Shen,
  C.}, \bibinfo{author}{Reid, I.D.}, \bibinfo{year}{2017}.
\newblock \bibinfo{title}{Refinenet: Multi-path refinement networks for
  high-resolution semantic segmentation.}, in: \bibinfo{booktitle}{Cvpr},
  p.~\bibinfo{pages}{5}.
\bibitem[{Long et~al.(2015)Long, Shelhamer and Darrell}]{long2015fully}
\bibinfo{author}{Long, J.}, \bibinfo{author}{Shelhamer, E.},
  \bibinfo{author}{Darrell, T.}, \bibinfo{year}{2015}.
\newblock \bibinfo{title}{Fully convolutional networks for semantic
  segmentation}, in: \bibinfo{booktitle}{Proceedings of the IEEE Conference on
  Computer Vision and Pattern Recognition}, pp. \bibinfo{pages}{3431--3440}.
\bibitem[{McIntosh et~al.(2020)McIntosh, Duarte, Rawat and
  Shah}]{mcIntosh2020vt}
\bibinfo{author}{McIntosh, B.}, \bibinfo{author}{Duarte, K.},
  \bibinfo{author}{Rawat, Y.S.}, \bibinfo{author}{Shah, M.},
  \bibinfo{year}{2020}.
\newblock \bibinfo{title}{Visual-textual capsule routing for text-based video
  segmentation}, in: \bibinfo{booktitle}{Proceedings of the IEEE Conference on
  Computer Vision and Pattern Recognition}, pp. \bibinfo{pages}{9942--9951}.
\bibitem[{Mehta et~al.(2018)Mehta, Mercan, Bartlett, Weave, Elmore and
  Shapiro}]{mehta2018net}
\bibinfo{author}{Mehta, S.}, \bibinfo{author}{Mercan, E.},
  \bibinfo{author}{Bartlett, J.}, \bibinfo{author}{Weave, D.},
  \bibinfo{author}{Elmore, J.G.}, \bibinfo{author}{Shapiro, L.},
  \bibinfo{year}{2018}.
\newblock \bibinfo{title}{Y-net: Joint segmentation and classification for
  diagnosis of breast biopsy images}.
\newblock \bibinfo{journal}{arXiv preprint arXiv:1806.01313} .
\bibitem[{Mobiny et~al.(2019)Mobiny, Lu, Nguyen, Roysam and
  Varadarajan}]{mobiny2019automated}
\bibinfo{author}{Mobiny, A.}, \bibinfo{author}{Lu, H.},
  \bibinfo{author}{Nguyen, H.V.}, \bibinfo{author}{Roysam, B.},
  \bibinfo{author}{Varadarajan, N.}, \bibinfo{year}{2019}.
\newblock \bibinfo{title}{Automated classification of apoptosis in phase
  contrast microscopy using capsule network}.
\newblock \bibinfo{journal}{IEEE transactions on medical imaging}
  \bibinfo{volume}{39}, \bibinfo{pages}{1--10}.
\bibitem[{Noh et~al.(2015)Noh, Hong and Han}]{noh2015deconv}
\bibinfo{author}{Noh, H.}, \bibinfo{author}{Hong, S.}, \bibinfo{author}{Han,
  B.}, \bibinfo{year}{2015}.
\newblock \bibinfo{title}{Learning deconvolution network for semantic
  segmentation}, in: \bibinfo{booktitle}{Proceedings of the IEEE International
  Conference on Computer Vision}, pp. \bibinfo{pages}{1520--1528}.
\bibitem[{Pal et~al.(2018)Pal, Chaturvedi, Garain, Chandra, Chatterjee and
  Senapati}]{pal2018capsdemm}
\bibinfo{author}{Pal, A.}, \bibinfo{author}{Chaturvedi, A.},
  \bibinfo{author}{Garain, U.}, \bibinfo{author}{Chandra, A.},
  \bibinfo{author}{Chatterjee, R.}, \bibinfo{author}{Senapati, S.},
  \bibinfo{year}{2018}.
\newblock \bibinfo{title}{Capsdemm: Capsule network for detection of munro’s
  microabscess in skin biopsy images}, in: \bibinfo{booktitle}{International
  Conference on Medical Image Computing and Computer-Assisted Intervention},
  \bibinfo{organization}{Springer}. pp. \bibinfo{pages}{389--397}.
\bibitem[{Ravishankar et~al.(2017)Ravishankar, Venkataramani, Thiruvenkadam,
  Sudhakar and Vaidya}]{ravishankar2017shape}
\bibinfo{author}{Ravishankar, H.}, \bibinfo{author}{Venkataramani, R.},
  \bibinfo{author}{Thiruvenkadam, S.}, \bibinfo{author}{Sudhakar, P.},
  \bibinfo{author}{Vaidya, V.}, \bibinfo{year}{2017}.
\newblock \bibinfo{title}{Learning and incorporating shape models for semantic
  segmentation}, in: \bibinfo{booktitle}{International Conference on Medical
  Image Computing and Computer-Assisted Intervention},
  \bibinfo{organization}{Springer}. pp. \bibinfo{pages}{203--211}.
\bibitem[{Ronneberger et~al.(2015)}]{ronneberger2015unet}
\bibinfo{author}{Ronneberger, O.}, et~al., \bibinfo{year}{2015}.
\newblock \bibinfo{title}{U-net: Convolutional networks for biomedical image
  segmentation}, in: \bibinfo{booktitle}{MICCAI15},
  \bibinfo{organization}{Springer}. pp. \bibinfo{pages}{234--241}.
\bibitem[{Sabour et~al.(2017)Sabour, Frosst and Hinton}]{hinton2017nips}
\bibinfo{author}{Sabour, S.}, \bibinfo{author}{Frosst, N.},
  \bibinfo{author}{Hinton, G.E.}, \bibinfo{year}{2017}.
\newblock \bibinfo{title}{Dynamic routing between capsules}, in:
  \bibinfo{booktitle}{Advances in Neural Information Processing Systems}, pp.
  \bibinfo{pages}{3856--3866}.
\bibitem[{Simonyan and Zisserman(2014)}]{simonyan2014vgg}
\bibinfo{author}{Simonyan, K.}, \bibinfo{author}{Zisserman, A.},
  \bibinfo{year}{2014}.
\newblock \bibinfo{title}{Very deep convolutional networks for large-scale
  image recognition}.
\newblock \bibinfo{journal}{arXiv preprint arXiv:1409.1556} .
\bibitem[{Song et~al.(2015)Song, Wu, Sun, Bahrami, Li and
  Shen}]{song2015progressive}
\bibinfo{author}{Song, Y.}, \bibinfo{author}{Wu, G.}, \bibinfo{author}{Sun,
  Q.}, \bibinfo{author}{Bahrami, K.}, \bibinfo{author}{Li, C.},
  \bibinfo{author}{Shen, D.}, \bibinfo{year}{2015}.
\newblock \bibinfo{title}{Progressive label fusion framework for multi-atlas
  segmentation by dictionary evolution}, in: \bibinfo{booktitle}{International
  Conference on Medical Image Computing and Computer-Assisted Intervention},
  \bibinfo{organization}{Springer}. pp. \bibinfo{pages}{190--197}.
\bibitem[{Srivastava et~al.(2014)Srivastava, Hinton, Krizhevsky, Sutskever and
  Salakhutdinov}]{srivastava2014dropout}
\bibinfo{author}{Srivastava, N.}, \bibinfo{author}{Hinton, G.},
  \bibinfo{author}{Krizhevsky, A.}, \bibinfo{author}{Sutskever, I.},
  \bibinfo{author}{Salakhutdinov, R.}, \bibinfo{year}{2014}.
\newblock \bibinfo{title}{Dropout: a simple way to prevent neural networks from
  overfitting}.
\newblock \bibinfo{journal}{The Journal of Machine Learning Research}
  \bibinfo{volume}{15}, \bibinfo{pages}{1929--1958}.
\bibitem[{Sun et~al.(2019)Sun, Zhao, Jiang, Cheng, Xiao, Liu, Mu, Wang, Liu and
  Wang}]{sun2019high}
\bibinfo{author}{Sun, K.}, \bibinfo{author}{Zhao, Y.}, \bibinfo{author}{Jiang,
  B.}, \bibinfo{author}{Cheng, T.}, \bibinfo{author}{Xiao, B.},
  \bibinfo{author}{Liu, D.}, \bibinfo{author}{Mu, Y.}, \bibinfo{author}{Wang,
  X.}, \bibinfo{author}{Liu, W.}, \bibinfo{author}{Wang, J.},
  \bibinfo{year}{2019}.
\newblock \bibinfo{title}{High-resolution representations for labeling pixels
  and regions}.
\newblock \bibinfo{journal}{arXiv preprint arXiv:1904.04514} .
\bibitem[{Tong et~al.(2013)Tong, Wolz, Coup{\'e}, Hajnal, Rueckert, Initiative
  et~al.}]{tong2013segmentation}
\bibinfo{author}{Tong, T.}, \bibinfo{author}{Wolz, R.},
  \bibinfo{author}{Coup{\'e}, P.}, \bibinfo{author}{Hajnal, J.V.},
  \bibinfo{author}{Rueckert, D.}, \bibinfo{author}{Initiative, A.D.N.}, et~al.,
  \bibinfo{year}{2013}.
\newblock \bibinfo{title}{Segmentation of mr images via discriminative
  dictionary learning and sparse coding: Application to hippocampus labeling}.
\newblock \bibinfo{journal}{NeuroImage} \bibinfo{volume}{76},
  \bibinfo{pages}{11--23}.
\bibitem[{Wu et~al.(2015)Wu, Kim, Sanroma, Wang, Munsell, Shen, Initiative
  et~al.}]{wu2015hierarchical}
\bibinfo{author}{Wu, G.}, \bibinfo{author}{Kim, M.}, \bibinfo{author}{Sanroma,
  G.}, \bibinfo{author}{Wang, Q.}, \bibinfo{author}{Munsell, B.C.},
  \bibinfo{author}{Shen, D.}, \bibinfo{author}{Initiative, A.D.N.}, et~al.,
  \bibinfo{year}{2015}.
\newblock \bibinfo{title}{Hierarchical multi-atlas label fusion with
  multi-scale feature representation and label-specific patch partition}.
\newblock \bibinfo{journal}{NeuroImage} \bibinfo{volume}{106},
  \bibinfo{pages}{34--46}.
\bibitem[{Yu and Koltun(2015)}]{yu2015dilate}
\bibinfo{author}{Yu, F.}, \bibinfo{author}{Koltun, V.}, \bibinfo{year}{2015}.
\newblock \bibinfo{title}{Multi-scale context aggregation by dilated
  convolutions}.
\newblock \bibinfo{journal}{arXiv preprint arXiv:1511.07122} .
\bibitem[{Yu et~al.(2017)Yu, Koltun and Funkhouser}]{yu2017dilated}
\bibinfo{author}{Yu, F.}, \bibinfo{author}{Koltun, V.},
  \bibinfo{author}{Funkhouser, T.}, \bibinfo{year}{2017}.
\newblock \bibinfo{title}{Dilated residual networks}, in:
  \bibinfo{booktitle}{Computer Vision and Pattern Recognition}.
\bibitem[{Zemel et~al.(1990)Zemel, Mozer and Hinton}]{zemel1990traffic}
\bibinfo{author}{Zemel, R.S.}, \bibinfo{author}{Mozer, M.C.},
  \bibinfo{author}{Hinton, G.E.}, \bibinfo{year}{1990}.
\newblock \bibinfo{title}{Traffic: Recognizing objects using hierarchical
  reference frame transformations}, in: \bibinfo{booktitle}{Advances in neural
  information processing systems}, pp. \bibinfo{pages}{266--273}.
\bibitem[{Zhang et~al.(2018a)Zhang, Zhou and Wu}]{zhang2018fast}
\bibinfo{author}{Zhang, S.}, \bibinfo{author}{Zhou, Q.}, \bibinfo{author}{Wu,
  X.}, \bibinfo{year}{2018}a.
\newblock \bibinfo{title}{Fast dynamic routing based on weighted kernel density
  estimation}, in: \bibinfo{booktitle}{International Symposium on Artificial
  Intelligence and Robotics}, pp. \bibinfo{pages}{301--309}.
\bibitem[{Zhang et~al.(2018b)Zhang, Zhang, Peng, Xue and Sun}]{zhang2018exfuse}
\bibinfo{author}{Zhang, Z.}, \bibinfo{author}{Zhang, X.},
  \bibinfo{author}{Peng, C.}, \bibinfo{author}{Xue, X.}, \bibinfo{author}{Sun,
  J.}, \bibinfo{year}{2018}b.
\newblock \bibinfo{title}{Exfuse: Enhancing feature fusion for semantic
  segmentation}, in: \bibinfo{booktitle}{The European Conference on Computer
  Vision (ECCV)}.
\bibitem[{Zhao et~al.(2017)Zhao, Shi, Qi, Wang and Jia}]{zhao2017pyramid}
\bibinfo{author}{Zhao, H.}, \bibinfo{author}{Shi, J.}, \bibinfo{author}{Qi,
  X.}, \bibinfo{author}{Wang, X.}, \bibinfo{author}{Jia, J.},
  \bibinfo{year}{2017}.
\newblock \bibinfo{title}{Pyramid scene parsing network}, in:
  \bibinfo{booktitle}{IEEE Conf. on Computer Vision and Pattern Recognition
  (CVPR)}, pp. \bibinfo{pages}{2881--2890}.
\bibitem[{Zhao et~al.(2018)Zhao, Ye, Yang, Lei, Zhang and
  Zhao}]{zhao2018investigating}
\bibinfo{author}{Zhao, W.}, \bibinfo{author}{Ye, J.}, \bibinfo{author}{Yang,
  M.}, \bibinfo{author}{Lei, Z.}, \bibinfo{author}{Zhang, S.},
  \bibinfo{author}{Zhao, Z.}, \bibinfo{year}{2018}.
\newblock \bibinfo{title}{Investigating capsule networks with dynamic routing
  for text classification}.
\newblock \bibinfo{journal}{arXiv preprint arXiv:1804.00538} .
\bibitem[{Zhou et~al.(2019)Zhou, Ji, Su, Sun and Chen}]{zhou2019dynamic}
\bibinfo{author}{Zhou, Y.}, \bibinfo{author}{Ji, R.}, \bibinfo{author}{Su, J.},
  \bibinfo{author}{Sun, X.}, \bibinfo{author}{Chen, W.}, \bibinfo{year}{2019}.
\newblock \bibinfo{title}{Dynamic capsule attention for visual question
  answering}, in: \bibinfo{booktitle}{Proceedings of the AAAI Conference on
  Artificial Intelligence}, pp. \bibinfo{pages}{9324--9331}.
\bibitem[{Zoph et~al.(2020)Zoph, Ghiasi, Lin, Cui, Liu, Cubuk and
  Le}]{zoph2020rethinking}
\bibinfo{author}{Zoph, B.}, \bibinfo{author}{Ghiasi, G.}, \bibinfo{author}{Lin,
  T.Y.}, \bibinfo{author}{Cui, Y.}, \bibinfo{author}{Liu, H.},
  \bibinfo{author}{Cubuk, E.D.}, \bibinfo{author}{Le, Q.V.},
  \bibinfo{year}{2020}.
\newblock \bibinfo{title}{Rethinking pre-training and self-training}.
\newblock \bibinfo{journal}{arXiv preprint arXiv:2006.06882} .

\end{thebibliography}
\bibliographystyle{elsarticle-harv}

\newpage
\setcounter{secnumdepth}{0}
\section{Appendix A. Layer Configurations}  \label{sec:appxA}

\begin{center}
\scalebox{0.9}{
  \begin{tabu}{c|  c | c }
    \hline
    \hline
    \textbf{Layer Name} & \textbf{Configuration} & \textbf{Output Size}\\

    \hline
    input & -- & 24$\times$56$\times$1  \\
    \hline
    conv1 & kernel:3$\times$3, stride:1, padding:1 & 24$\times$56$\times$32 \\
    pool1 & kernel:2$\times$2, stride:1, padding:0 & 12$\times$28$\times$32 \\
    \hline
    conv2 & kernel:3$\times$3, stride:1, padding:1 & 12$\times$28$\times$64 \\
    pool2 & kernel:2$\times$2, stride:1, padding:0 & 6$\times$14$\times$64 \\
    \hline
    conv3 & kernel:3$\times$3, stride:1, padding:1 & 6$\times$14$\times$128 \\
    pool3 & kernel:2$\times$2, stride:1, padding:0 & 3$\times$7$\times$128 \\
    pool3\_dropout & dropout rate: 0.5 & 3$\times$7$\times$128\\
    \hline
    \hline
    deconv1 & kernel:2$\times$2, stride:2, padding:0 & 6$\times$14$\times$128 \\
    deconv1\_conv & kernel:3$\times$3, stride:1, padding:1 & 6$\times$14$\times$128 \\
    concat1 & concatenation of deconv1 with pool2 & 6$\times$14$\times$192 \\
   concat1\_dropout & dropout rate: 0.5 & 6$\times$14$\times$192\\
    concat1\_conv & kernel:3$\times$3, stride:1, padding:1 & 6$\times$14$\times$128 \\
    \hline
    deconv2 & kernel:2$\times$2, stride:2, padding:0 & 12$\times$28$\times$128 \\
    deconv2\_conv & kernel:3$\times$3, stride:1, padding:1 & 12$\times$28$\times$128 \\
    concat2 & concatenation of deconv2 with pool1 & 12$\times$28$\times$160 \\
    concat2\_dropout & dropout rate: 0.5 & 12$\times$28$\times$160\\
    deconv2\_conv & kernel:3$\times$3, stride:1, padding:1 & 12$\times$28$\times$128 \\
    \hline
    deconv3 & kernel:2$\times$2, stride:2, padding:0 & 24$\times$56$\times$128 \\
    deconv3\_conv & kernel:3$\times$3, stride:1, padding:1 & 24$\times$56$\times$128 \\
   \hline
   class\_conv & kernel:3$\times$3, stride:1, padding:1 & 24$\times$56$\times$2 \\
  \end{tabu}}
  \captionof{table}{Layer configuration of the baseline U-Net model with its smallest feature map 3$\times$7.}
\label{tab:baseline}

\end{center}

\begin{center}
\scalebox{0.9}{
  \begin{tabu}{c|  c | c }
    \hline
    \hline
    \textbf{Layer Name} & \textbf{Configuration} & \textbf{Output Size}\\

    \hline
    input & -- & 24$\times$56$\times$1  \\
    \hline
    conv1 & kernel:3$\times$3, stride:1, padding:1 & 24$\times$56$\times$32 \\
    pool1 & kernel:2$\times$2, stride:1, padding:0 & 12$\times$28$\times$32 \\
    conv2 & kernel:3$\times$5, stride:1, padding:0 & 8$\times$24$\times$64 \\
    conv2\_dropout & dropout rate: 0.5 & 8$\times$24$\times$64\\
    \hline
    \hline
    primary capsules & \makecell{modified convolution layer \\ kernel:5$\times$5, stride:1, padding:0,  \\capsules:32, capsule size:8} &  4$\times$20$\times$(32$\times$8) \\
    \hline
    class capsules &\makecell{capsule layers with 1$\times$1 kernel, \\capsules:2, capsule size:16} & 2$\times$16\\
    \hline
    traceback & traceback the probability map & 4$\times$20$\times$2 \\
    \hline
    conv3 & kernel:3$\times$3, stride:1, padding:1 & 4$\times$20$\times$256 \\
    \hline
    \hline
    deconv1 & kernel:5$\times$5, stride:1, padding:0 & 8$\times$24$\times$128 \\
    concat1 & concatenation of deconv1 with conv2 & 8$\times$24$\times$192 \\
    concat1\_dropout & dropout rate: 0.5 & 8$\times$24$\times$192\\
    deconv1\_conv & kernel:3$\times$3, stride:1, padding:1 & 8$\times$24$\times$128 \\
    \hline
    deconv2 & kernel:5$\times$5, stride:1, padding:0 & 12$\times$28$\times$128 \\
    concat2 & concatenation of deconv2 with pool1 & 12$\times$28$\times$160 \\
    deconv2\_conv & kernel:3$\times$3, stride:1, padding:1 & 12$\times$28$\times$128 \\
    \hline
    deconv3 & kernel:4$\times$4, stride:2, padding:0 & 26$\times$58$\times$128 \\
    \hline
    class\_conv & kernel:3$\times$3, stride:1, padding:1 & 24$\times$56$\times$2 \\
   \hline
  \end{tabu}}
  \captionof{table}{Layer configuration of TraceCaps with its traceback size 4$\times$20.}
\label{tab:layerconfig}
\end{center}

%
%

%
%
%
\end{document}